\documentclass[10pt,twocolumn,letterpaper]{article}

\usepackage{wacv}              

\usepackage{graphicx}
\usepackage{amsmath}
\usepackage{amssymb}
\usepackage{booktabs}
\usepackage{xcolor}

\usepackage[pagebackref,breaklinks,colorlinks]{hyperref}
\usepackage{algorithm}
\usepackage{algorithmic}

\usepackage[capitalize]{cleveref}
\crefname{section}{Sec.}{Secs.}
\Crefname{section}{Section}{Sections}
\Crefname{table}{Table}{Tables}
\crefname{table}{Tab.}{Tabs.}

\def\wacvPaperID{2337} 
\def\confName{WACV}
\def\confYear{2025}

\begin{document}

\title{Leveraging CLIP Encoder for Multimodal Emotion Recognition}

\author{Yehun Song\\
Agency for Defense Development\\
Daejeon, Republic of Korea\\
{\tt\small kidswave@add.re.kr}
\and
Sunyoung Cho\\
Sookmyung Women's University\\
Seoul, Republic of Korea\\
{\tt\small sycho22@sookmyung.ac.kr}
}
\maketitle
\begin{abstract}
Multimodal emotion recognition (MER) aims to identify human emotions by combining data from various modalities such as language, audio, and vision. Despite the recent advances of MER approaches, the limitations in obtaining extensive datasets impede the improvement of performance. To mitigate this issue, we leverage a Contrastive Language-Image Pre-training (CLIP)-based architecture and its semantic knowledge from massive datasets that aims to enhance the discriminative multimodal representation. 
We propose a label encoder-guided MER framework based on CLIP (MER-CLIP) to learn emotion-related representations across modalities.
Our approach introduces a label encoder that treats labels as text embeddings to incorporate their semantic information, leading to the learning of more representative emotional features. 
To further exploit label semantics, we devise a cross-modal decoder that aligns each modality to a shared embedding space by sequentially fusing modality features based on emotion-related input from the label encoder.
Finally, the label encoder-guided prediction enables generalization across diverse labels by embedding their semantic information as well as word labels.
Experimental results show that our method outperforms the state-of-the-art MER methods on the benchmark datasets, CMU-MOSI and CMU-MOSEI.
\end{abstract}

\section{Introduction}
\label{sec:intro}
Multimodal emotion recognition (MER) is one of the representative problems in multimodal learning, aiming to perceive human emotions from various modalities, including language, audio, and vision~\cite{tadas19, trisha20}. Such rich multimodal information enables the understanding of human behaviors and intents~\cite{manning14} by reducing emotional ambiguities from one modality~\cite{alswaidan20, hu18, wang21} or two modalities~\cite{ngiam11, sheikh18} through the utilization of their complementarity. 

There has been significant progress in MER with algorithms that aim to learn discriminative representations and fuse heterogeneous information from multiple sources. Most of them have used benchmark multimodal datasets, such as IEMOCAP~\cite{iemocap}, CMU-MOSI~\cite{mosi}, CMU-MOSEI~\cite{mosei}, MELD~\cite{meld}, and so on. However, these datasets containing three-modality data are typically smaller than those with single- or dual-modality data. For example, CMU-MOSEI, one of the large datasets in MER, contains 23,259 video clips involving three modalities, while AffectNet~\cite{ali19} contains 450,000 images for facial expression recognition. For MER, obtaining a large-scale dataset with tri-modal data is extremely difficult due to label uncertainties as well as a lack of valid data. The dataset needs to collect time-series data involving both facial expressions and speech of the speaker across various emotional states. Label uncertainties caused by the subjectivity of annotators within certain emotion classes often result in inconsistent and incorrect labels~\cite{wang20}. Therefore, multimodal learning with limited and uncertain datasets inherently imposes limitations on performance improvement, regardless of improvements in algorithmic performance. 

Recently, multimodal models~\cite{radford21, guzhov22, wu22, ruan23} pretrained from large-scale datasets have achieved great success in a variety of tasks. The most popular Vision-Language model is Contrastive Language-Image Pre-training (CLIP)~\cite{radford21}. CLIP enables to encode rich semantic information by contrastively training with a large number of image-text pairs collected from the internet, resulting in high zero-shot performance in image classification, image retrieval, and generative models. Especially, the representation learning of CLIP has been successfully exploited for facial expression recognition~\cite{DBLP:conf/bmvc/ZhaoP23, foteinopoulou24, chen24}, indicating a potential for the CLIP learning framework to emotion recognition task. 
However, since such representation learning uses a single visual modality, the current scheme is unable to reflect multiple modalities for MER.

To address these issues, we propose a label encoder-guided MER framework based on CLIP (MER-CLIP), which aims to capture richer semantic relationships among modalities from large-scale pretrained models. Instead of employing conventional modality features~\cite{pennington14, tadas16, degottex14} in previous works, we leverage the pretrained weights of the CLIP architecture for modality encoders. To consistently learn cross-modal correlations exploiting the embedding space of CLIP text encoder, we devise a label encoder-guided prediction mechanism that utilizes emotion-aligned features to extract multimodal representations. Consequently, the same unified framework enables the learning of abundant multimodal representations in a more consistent and efficient manner.

Our framework adapts the CLIP text encoder to fuse and align the feature spaces of each modality into a joint multimodal representation with embedding vectors, which are associated with emotion labels. To capture cross-modal correlations, we introduce a cross-modal decoder that sequentially fuses features from different modalities. It reinforces the multimodal representations by considering modality interactions and reducing redundancy and discrepancies across different modalities. Emotion-related embeddings from the label encoder are used as queries for the cross-modal decoder to provide emotion-aligned representations across multiple modalities. The resulting multimodal representations are employed to predict emotion labels by calculating their cosine similarity with embedding vectors obtained from the label encoder. Differing from the use of a fully-connected output layer in previous works, our learning scheme based on the text encoder provides a more generalizable and adaptable method for MER, enabling the handling of diverse labels such as phrases, sentences, as well as words.

In summary, our main contributions are as follows:
\begin{itemize}
\item We propose a label encoder-guided multimodal representation learning framework (MER-CLIP) that exploits rich semantics from large-scale pretrained models for MER. We employ the same CLIP architecture for each modality encoder and learn the correlation across modalities via emotion-related embeddings from the cross-modal decoder and label encoder. Such framework provides a consistent and efficient training manner where the same unified multimodal encoder structure can be learned.
\item We present a cross-modal decoder that utilizes multimodal features extracted from data encoders as key-value pairs, while employs embeddings associated with the word of `Emotion' from the label encoder as queries. Our decoder provides emotion-aligned representations across multiple modalities for MER.
\item MER-CLIP employs the cosine similarity between multimodal representations and embedding vectors obtained from the label encoder (pretrained CLIP text encoder) for label prediction, rather than relying on a fully-connected output layer. This approach enhances generalizability and adaptability across diverse labels including phrases and sentences. 
\end{itemize}

\begin{figure*}
 \centering
 \includegraphics[height=8cm]{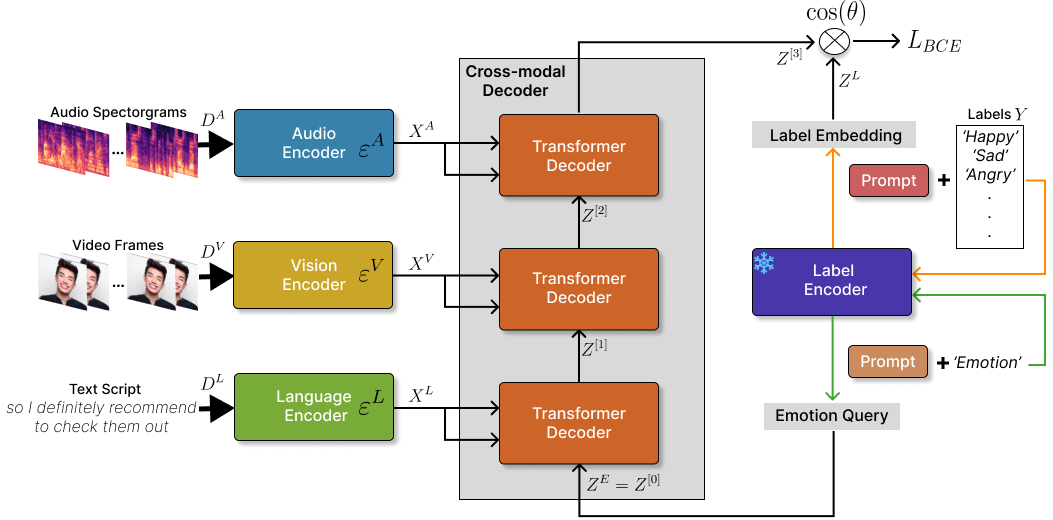}
 \caption{Overall structure of MER-CLIP. Given the input multimodal data $D^m$, MER-CLIP first encodes modality features $X^m$, via CLIP-based modality encoders $\varepsilon^m$ ($m \in \left \{V, L, A \right \}$). In cross-modal decoder, emotion-aligned multimodal representations $Z^{[k]}$ are extracted based on emotion query embeddings $Z^E$ from label encoder. Finally, the cosine similarity between the final representation $Z^{[3]}$ and the label embedding $Z^L$ is exploited for emotion prediction.}
 \label{fig:model1}
\end{figure*}

\section{Related Work}
\label{sec:formatting}
\subsection{Multimodal Emotion Recognition}
Multimodal emotion recognition(MER) aims at identifying human emotions by using multiple modalities including vision, audio, and language modalities. Depending on how to extract the contextual information from multi-source data, existing MER methods can be divided into two categories: fusion-based methods~\cite{ngiam11, zadeh17, hazarika20} and alignment-based methods~\cite{tsai19, zhang22, lv21}.

The fusion-based methods focus on generating discriminative multimodal representations by combining features of multiple modalities. Early works adopted simple feature concatenation~\cite{ngiam11} and inter-/intra-modality dynamics modeling via tensor/memory fusion network~\cite{zadeh17, zadeh18}. To reduce the modality gap, some works~\cite{hazarika20, tsai18, yang22} introduce to decompose multimodal features into common and specific information to learn comprehensive representation for multimodal fusion. However, these methods tend to occur information loss or corruption of modality characteristics during feature decoupling procedure.

The alignment-based methods aim to learn the reinforced representation by identifying the correlations between different modalities. MulT~\cite{tsai19} presented a cross-modal attention mechanism to learn the correlations from source modality to target modality and it leads to semantic implicit alignment. However, their approach performed independent modality reinforcement between modality pairs and does not consider the three-way correlations across all the involved modalities. To solve the pairwise fusion problem, TAILOR~\cite{zhang22} introduced a hierarchical cross-modal encoder that gradually considers interactions across modalities. PMR~\cite{lv21} proposed a message hub to reinforce the modality features by progressively exchanging information with each modality. In this work, we propose a new approach that explores correlations across all modalities by aligning multimodal features into the CLIP text encoder embedding space~\cite{radford21}.

\subsection{Large-scale pre-trained multimodal model}
Large-scale pre-trained multimodal model aims to encode rich semantics from massive datasets for multimodal learning. CLIP~\cite{radford21} is a representative Vision-Language model that is trained on a web-scale data of image and caption pairs through contrastive self-supervised objectives. Due to the great success of the CLIP, it has been applied to various vision-language tasks such as image-text pairing~\cite{christoph21}, image classification and retrieval~\cite{pmlr-v182-zhang22a}, image generation~\cite{rinon22}, and image captioning~\cite{pmlr-v162-fang22a}.

Some works~\cite{guzhov22, wu22, ruan23} have attempted to integrate audio modality into the CLIP framework. AudioCLIP~\cite{guzhov22} combines CLIP with the ESResNeXt audio encoder~\cite{guzhov21} and extracts the audio representation by contrastive learning across image, text and audio. Wav2CLIP~\cite{wu22} freezes the CLIP image encoder to contrastively learn the ResNet-18-based audio encoder for encoding all modalities to a shared embedding space. CLIP4VLA~\cite{ruan23} learns semantic representations for vision, audio and text modalities by contrastively training inter-modal and intra-modal correlations. The audio representation is enhanced by applying an audio augmentation and audio type token to distinguish between verbal and non-verbal information.

However, previous works mainly focus on learning the correlation across three modalities, applying them to the task of matching the most appropriate text description based on visual and audio information such as retrieval and captioning. Unlike commonly considered tasks, in the MER task, the text data comprises the content of the speaker's speech rather than explaining visual and audio information. Since all information from three modalities should be considered for emotion recognition, previous multimodal models cannot be directly applied to MER. Different from them, we aim to use the additional text encoder in the Vision-Language-Audio model to produce label embeddings that can be directly applied to general classification tasks as well as MER.

\section{MER-CLIP}
In this section, we provide a detailed description of our proposed MER-CLIP and its training scheme. 
The overall structure is illustrated in Fig.~\ref{fig:model1}.
It consists of three parts: modality encoders, cross-modal decoder, label encoder-guided prediction.
To encode rich semantics from large-scale pre-trained model, we leverage the pretrained weights of the CLIP for extracting modality features. 
The details of modality encoders are introduced in Sec.~\ref{sec:modality_encoder}. 
To obtain refined and reinforced multimodal representations, we learn correlations across modalities using a cross-modal decoder based on attention mechanism. Emotion-aligned representations are produced by exploiting emotion-related embeddings from the label encoder.
The detail of the cross-modal decoder is introduced in Sec.~\ref{sec:crossmodal_decoder}.
Finally, the cosine similarity between the embeddings from the final output of the cross-modal decoder and the label embeddings from the label encoder is exploited for MER, as introduced in Sec.~\ref{sec:label_encoder}. 
Our goal is to ensure that each modality is aligned and encoded with its corresponding emotion label.

\subsection{Modality encoders}
\label{sec:modality_encoder}

In this work, we consider three modalities, \ie, vision $(V)$, language $(L)$, audio $(A)$. 
Let $D^m$ denotes the input multimodal data, where $m \in \left \{V, L, A \right \}$ indicates a modality, and $Y$ be the emotion labels. 
Given a training dataset
$D=\left \{ (D_i^V, D_i^L, D_i^A), Y_i \right \}_{i=1}^{n}$, we exploit three CLIP encoders to extract modality features $X^m$. Below, we present the details of the modality encoders $\varepsilon^m$.

\subsubsection{Vision encoder}
We employ pretrained CLIP image encoder to extract vision modality features for the visual input $D^V$.
We first sample $T_V$ frames for each video $D_i^V$ in the temporal dimension to form the frame sequence \{$ f_i^j $\}$_{j=1}^{T_V}$.
Each frame is split into a sequence of non-overlapping patches and then appended with $\texttt{[CLS]}$ token.
The sequence of patches in each frame is then fed into the vision encoder to capture the spatial relationship between patches.
We use the final output $v_i^j$ of the class token as the visual representation of each frame $f_i^j$.
Finally, we obtain a frame-level visual representation $v_i=$\{$ v_i^j $\}$_{j=1}^{T_V}$ for each video $D_i^V$ and then acquire a global visual embedding $X_i^V$ by applying the average pooling of $v_i$ to learn temporal representation~\cite{rasheed23}.

\subsubsection{Language encoder}
We employ pretrained CLIP text encoder to extract language modality features for the language input $D^L$.
First, each language data $D_i^L$ is tokenized into a sequence of tokens with the addition of start and end tokens ($\texttt{[SOS]}$ and $\texttt{[EOS]}$), denoted by \{$ t_i^j $\}$_{j=1}^{T_L}$.
The sequence of tokens is fed into the text encoder, and word-level language representation $l_i=$\{$ l_i^j $\}$_{j=1}^{T_L}$ is obtained by collecting the output $l_i^j$ of each token $t_i^j$. 
We use the output of the $\texttt{[EOS]}$ token of $l_i$ as the global language representation $X_i^L$ for each language data $D_i^L$.

\subsubsection{Audio encoder}
Following previous works~\cite{wu22, ruan23}, we transform the audio waveform input $D^A$ into  spectrograms for use in the audio encoder.
The spectrogram is considered as a 2D image, leading to design the audio encoder with the same structure of the CLIP image encoder.
To convert each audio input $D_i^A$ into a spectrogram, we divide the audio into 2-second segments and extract 224-dimensional log Mel-filterbank features with window length of 64 ms and hop size of 32 ms for each segment.
For $\tau$-second audio, we obtain the frame sequence \{$ s_i^j $\}$_{j=1}^{T_A}$, where $s_i^j \in \mathbb{R}^{224 \times 224 \times 3}$ and $T_A=\left \lceil \frac{\tau}{2} \right \rceil$, and it is appended with $\texttt{[CLS]}$ token.
The final sequence of audio segment features $a_i=$\{$ a_i^j $\}$_{j=1}^{T_A}$ is obtained by using the final output $a_i^j$ of the class token of each frame $s_i^j$.
Finally, we obtain a global audio embedding $X_i^A$ by applying average pooling of $a_i$.

\subsection{Cross-modal decoder} 
\label{sec:crossmodal_decoder}

\begin{figure}[tb]
 \centering
 \includegraphics[height=6cm]{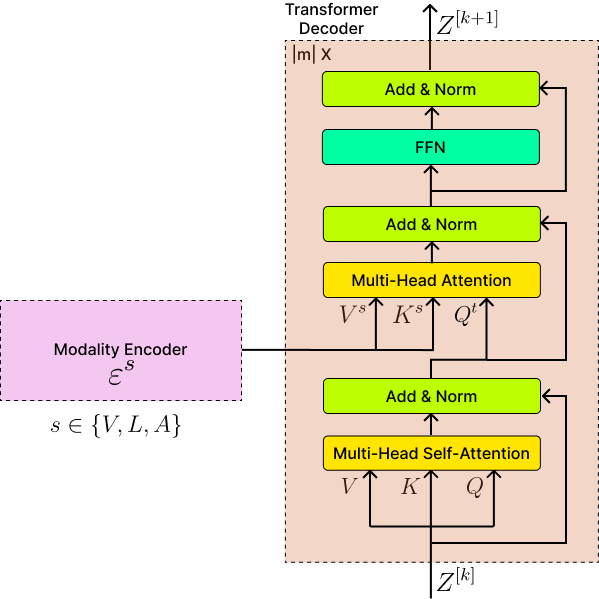}
 \caption{The structure of CMD. Given the emotion-related input $Z^{[k]}$, CMD learns to correlate representative elements across modalities using the multi-head attention layer. 
 In the layer, it learns correlations with the emotional features from the modality encoders $\varepsilon^s$. It receives key $K^s$ and value $V^s$ pair from the encoder, and query $Q^t$ from the self-attention layer. After that, it passes the feed-forward network and outputs the embeddings of $Z^{[k+1]}$. It is repeated across all three modalities.}
\label{fig:model2}
\end{figure}

To mitigate the feature distribution gap among different modalities, we utilize the cross-modal attention operation to enhance the multimodal representations. 
The typical cross-modal attention mechanism reinforces a target modality by incorporating information from a source modality through the learning of directional pairwise attention between components across modalities~\cite{tsai19}.
Due to independent pairwise fusion, it fails to exchange information across all modalities, resulting in information redundancy and an inability to exploit high-level features from the source modality.
Lv~\etal~\cite{lv21} overcome these issues by introducing a message hub to exchange information with each modality. 
However, it still fails to learn emotion-related correlations solely by using modality features as input to cross-modal attention operations.

Motivated by above observations, we investigate a cross-modal decoder (CMD) using multi-head attention mechanism, as shown in Fig.~\ref{fig:model2}.
It consists of $\left | m \right |$ layers to combine the feature representations of each modality as a source and emotion query from the label encoder as a target.
Let $X^s$ denotes the feature representations from the source modalities, where $s \in \left \{V, L, A \right \}$.
$X^t$ denotes the target representations from emotion query and enhanced representations $Z^{[k]}$, where $\left \{ k=0,\cdots, \left | m \right | \right \}$.
Our cross-modal attention unit is defined as: queries $Q^t=X^tW^{Q^t}$, keys $K^s=X^sW^{K^s}$, and values $V^s=X^sW^{V^s}$, where $W^{Q^t}, W^{K^s}, W^{V^s}$ are the learnable parameters.
The individual head of cross-modal attention can be defined as:

\begin{equation}
    Z_{s \rightarrow t}^{[k]}=\text{softmax}(\frac{Q^t {K^s}^{\top}}{\sqrt{d}})V^s
\end{equation}
where $Z_{s \rightarrow t}^{[k]}$ is the enhanced feature representation from $X^s$ to $X^t$.

The feature representation of each modality is sequentially added in the order of language $X^L$, visual $X^V$, and audio $X^A$.
The fusion order is determined by considering the degree of contribution of each modality to the final emotion prediction.
Generally, as language and visual modalities contribute the most, we opt to fuse them first to extract more representative emotional features.
Moreover, the initial input to CMD is the `Emotion Query' derived from a language embedding in the Label Encoder. 
Consequently, the CMD's embedding space is primarily shaped by the CLIP Text Encoder's language-based embeddings, making it more receptive to the language modality when introduced before the vision modality.
Therefore, we prioritize fusing the language modality first, then our model learns the cross-modal correlation with the fusion order $\left [ E, L, V, A \right ]$.
The final output $Z^{[\left | m \right |]}$ of CMD is used for label prediction.

\subsection{Label encoder-guided prediction}
\label{sec:label_encoder}

For label classification, the typical method involves employing a fully connected (FC) layer for the multimodal representations. 
However, this approach fails to utilize information from the labels themselves and is limited to exploiting the semantic relationships between labels. 
To address these issues, we propose a label encoder-guided prediction method.

We leverage a pre-trained CLIP text encoder as a label encoder (LE), utilizing the embeddings from the text encoder as labels.
For the initial input $Z^{[0]}$ to the cross-modal attention unit in CMD, we construct emotion query embeddings $Z^E$ using the output embedding of LE for the word `\textit{Emotion}', combined with a learnable prompt, to incorporate the global categorical information for the emotion.
The emotion-related initial input $Z^E$ enables CMD to learn to correlate emotion-related representative elements across multiple modalities.

Labels $Y$ are first wrapped in prompt $`` \left \{ \textit{learnable prompt} \right \} \left \{ \textit{label} \right \} "$ to generate text embeddings $Z^L$.
Similar to the processing of language input, each emotion label $Y_i \in Y$ is first tokenized as a sequence of tokens, followed by $\texttt{[SOS]}$ and $\texttt{[EOS]}$ tokens, denoted by \{$ l_i^j $\}$_{j=1}^{T_C}$.
After text encoding, the output of each token is collected as a word-level representation \{$ e_i^j $\}$_{j=1}^{T_C}$. 
We use the output of the $\texttt{[EOS]}$ token as the global textual representation of $l_i$, denoted by $e_i^g$.
Finally, the cosine similarity $sim(\cdot)$ between multimodal representation $Z_i$ and corresponding label embedding $e_i^g$ is maximized to fine-tune the MER-CLIP. 
For the sentiment analysis task, we use cross-entropy (CE) loss based on scaled pairwise $sim(\cdot)$.
For the emotion recognition task, where the dataset is multi-label, we apply binary cross-entropy (BCE) loss for the final classification: 

\begin{equation}
    L=-\frac{1}{N} \sum_{i=1}^N\left[Y_i \log \left(\hat{Y}_i\right)+\left(1-Y_i\right) \log \left(1-\hat{Y}_i\right)\right]
\end{equation}
where $Y_{i}$ is the ground-truth and $\hat{Y}_i$ is the predicted label by:

\begin{equation}
\hat{Y}_i=\text{sigmoid}(\left \| sim(Z_i, e_i^g) \right \|)
\end{equation}

In the training process of our model, the weight of LE is frozen to leverage the embeddings of the pretrained CLIP text encoder. 
The frozen embeddings are used to train the three modality encoders and CMD.
Each modality encoder is trained to extract discriminative modality features, while the CMD is aligned with the label embeddings of LE.
Note that LE is capable of processing any forms of text. 
To demonstrate the generalization capability of LE, experimental results based on different types of labels are described in Sec.~\ref{sec:generalization}.

\section{Experiments}
\label{sec:experiments}

\subsection{Datasets}
We evaluate MER-CLIP on CMU-MOSI~\cite{cmu_mosi} and CMU-MOSEI~\cite{mosei} datasets. 
CMU-MOSI is a multimodal sentiment analysis dataset consisting of 2,199 samples of short monologue video clips.
Following the predetermined setting, 1,284, 229, and 686 samples are used as training, validation and testing set.
CMU-MOSEI contains 22,856 samples of video clips from 1,000 distinct YouTube speakers for sentiment and emotion analysis.
Following the predetermined setting, 16,326, 1,871, and 4,659 samples are used for training, validation, and testing, respectively.

For both CMU-MOSI and CMU-MOSEI, each sample is annotated with a sentiment score from -3 to 3, including \emph{strongly negative}, \emph{negative}, \emph{weakly negative}, \emph{neutral}, \emph{weakly positive}, \emph{positive}, and \emph{strongly positive}.
Each sample in CMU-MOSEI is also annotated with multiple labels from 6 emotion categories, including \emph{happy}, \emph{sad}, \emph{angry}, \emph{surprise}, \emph{disgust}, and \emph{fear} for emotion recognition.
Each sample consists of three modalities: vision, language, and audio.
Since we require raw data to directly extract features from the CLIP encoder, we generate the utterance-level data from the raw CMU-MOSI and CMU-MOSEI datasets. 
This ensures a perfect match with the preprocessed data~\footnote{\url{https://github.com/CMU-MultiComp-Lab/CMU-MultimodalSDK}} in terms of the number of samples and labels, enabling fair comparison with other methods based on preprocessed data.
The experiments are conducted on the word-unaligned settings.

We employ four evaluation metrics to measure the performance of multi-label emotion classification~\cite{zhang14, zhang22}: Accuracy, Micro-F1, Precision, and Recall.
To evaluate the performance of sentiment analysis, we use two metrics: binary class accuracy ($\text{ACC}_{2}$: \emph{positive}/\emph{negative} sentiments) and F1 score.
Since sentiment labels are typically more ambiguous than emotion labels, it is challenging to fully leverage the semantic capability of the CLIP encoder.
Therefore, we evaluate the performance for two specific types of sentiment (\emph{positive} and \emph{negative}) that are more certain. 
Note that larger values indicate better performance.

\subsection{Baselines}

We compare MER-CLIP with the following current state-of-the-art MER methods.
For the emotion recognition task, we compare with multimodal multi-label methods, including DFG~\cite{mosei}, RAVEN~\cite{wang2019words}, MulT~\cite{tsai19}, SIMM~\cite{wu2019multi}, MISA~\cite{hazarika20}, HHMPN~\cite{zhang2021multi}, and TAILOR~\cite{zhang22}.
For the sentiment analysis task, we compare with multimodal methods, including Early Fusion LSTM (EF-LSTM), Late Fusion LSTM (LF-LSTM), RAVEN~\cite{wang2019words}, MCTN~\cite{pham2019aaai}, MulT~\cite{tsai19}, PMR~\cite{lv21}, MICA~\cite{Liang_2021_ICCV}, and DMD~\cite{li23decoupled}.

\subsection{Implementation details}

For the modality encoders, we initialize the vision and audio encoder  using the pretrained weights of the image encoder in CLIP, and the language encoder using the pretrained weights of the text encoder in CLIP.
We use ViT-B/32 backbone for CLIP.
We set the batch size to 16 (8 for MOSI) and trained our model for 20 epochs.
All parameters in our model are optimized by AdamW~\cite{loshchilov2017} with a learning rate of $8\textit{e}^{-6}$ ($2.2\textit{e}^{-5}$ for MOSI) and weight decay of 0.001. 
For the visual and audio encoders, we use the regularly selected 12 frames. 
We use 8 context vectors to construct the prompts for label embeddings and emotion queries.
For CMD, we use three hidden layers of size 512 with 8 heads.
For the initial input to CMD, we use query embeddings generated from LE using the word `\textit{Emotion}' for emotion recognition and the word `\textit{Sentiment}' for sentiment analysis.
Please see Appendix for more experimental results regarding various emotion-related words. 

\subsection{Comparison with state-of-the-arts}

We evaluate MER-CLIP under the unaligned setting, as our data encoders receives unaligned multimodal sequences, which is more challenging than the word-aligned setting.
Table~\ref{tab:cmu_result} shows the comparison for multimodal emotion recognition on CMU-MOSEI dataset.
MER-CLIP achieves state-of-the-art results on all evaluation metrics except precision.
DFG~\cite{mosei}, RAVEN~\cite{wang2019words}, MulT~\cite{tsai19}, HHMPN~\cite{zhang2021multi}, and TAILOR~\cite{zhang22} perform better than MER-CLIP on the precision metric, whereas MER-CLIP outperforms them on the recall metric. 
In general, there is a trade-off between precision and recall, making accuracy and micro-F1 more important for evaluation. 
MER-CLIP outperforms the previous multimodal multi-label methods on both accuracy and micro-F1, indicating the feasibility of our multimodal representation learning framework.


\begin{table}[]
\setlength{\tabcolsep}{0pt}
\small
\caption{Comparison for multimodal emotion recognition on CMU-MOSEI dataset.}
\label{tab:cmu_result}
\centering
\begin{tabular}{l|cccc}
\hline
\multicolumn{1}{c|}{Methods} & Accuracy(\%) & Precision(\%) & Recall(\%) & Micro-F1(\%) \\ \hline \hline
DFG~\cite{mosei}             & 38.6    & 53.4     & 45.6  & 49.4    \\
RAVEN~\cite{wang2019words}   & 40.3    & 63.3     & 42.9  & 51.1    \\
MulT~\cite{tsai19}           & 42.3    & 63.6     & 44.5  & 52.3    \\
SIMM~\cite{wu2019multi}      & 41.8    & 48.2     & 48.6  & 48.4    \\
MISA~\cite{hazarika20}       & 39.8    & 37.1     & 57.1  & 45.0    \\
HHMPN~\cite{zhang2021multi}  & 43.4    & 59.1     & 47.6  & 52.8    \\
TAILOR~\cite{zhang22}        & 46.0    & \textbf{63.9}     & 45.2  & 52.9    \\ \hline
MER-CLIP (Ours)               & \textbf{49.3}    & 53.1     & \textbf{63.4}  & \textbf{57.8}    \\ \hline
\end{tabular}
\end{table}

\begin{table}[]
\small
\caption{Comparison for multimodal sentiment analysis on CMU-MOSI and CMU-MOSEI datasets.}
\label{tab:cmu_result2}
\centering
\begin{tabular}{l|cccc}
\hline
 & \multicolumn{2}{c}{CMU-MOSI} & \multicolumn{2}{c}{CMU-MOSEI} \\

Methods & $\text{ACC}_{2}$(\%)         & F1(\%)         & $\text{ACC}_{2}$(\%)          & F1(\%)         \\ \hline \hline
EF-LSTM                      & 73.6    & 74.5   & 76.1 & 75.9 \\
LF-LSTM                      & 77.6    & 77.8   & 77.5 & 78.2 \\
RAVEN~\cite{wang2019words}   & 72.7    & 73.1   & 75.4 & 75.7 \\
MCTN~\cite{pham2019aaai}     & 75.9    & 76.4   & 79.3 & 79.7 \\
MulT~\cite{tsai19}           & 81.1    & 81.0   & 81.6 & 81.6 \\ 
PMR~\cite{lv21}              & 82.4    & 82.1   & 83.1 & 82.8 \\ 
MICA~\cite{Liang_2021_ICCV}  & 82.6    & 82.7   & 83.7 & 83.3 \\
DMD~\cite{li23decoupled}     & 83.5    & 83.5  & 84.8 & 84.7  \\ \hline
MER-CLIP (Ours)               & \textbf{84.0}    & \textbf{84.0} & \textbf{85.3} & \textbf{85.1} \\ \hline
\end{tabular}
\end{table}

Table~\ref{tab:cmu_result2} illustrates the comparison for multimodal sentiment analysis on CMU-MOSI and CMU-MOSEI datasets. 
Our MER-CLIP obtains superior MER accuracy than other MER methods including simple fusion-based methods (EF-LSTM, LF-LSTM) and state-of-the-art methods including DMD~\cite{li23decoupled} on both datasets. 
Compared with cross-modal interaction learning-based methods~\cite{tsai19, Liang_2021_ICCV, lv21}, our MER-CLIP achieves more significant improvements, indicating the feasibility of the cross-modal correlation learning based on rich modality features.

\subsection{Ablation study}
We perform ablation studies on the key components of MER-CLIP in the sentiment analysis task, including CMD and LE. The results are illustrated in Table~\ref{tab:modality_orders} and Table~\ref{tab:ablations}.

\subsubsection{Impact of inputs on CMD}
To validate the necessity of employing multiple modalities, we first compare the performance of uni-modality, bi-modality, and tri-modality inputs in CMD for sentiment analysis using CMU-MOSEI dataset, as shown in Table~\ref{tab:modality_orders}.
Among the three modalities, we observe that our method based on language modality features outperforms the multimodal transformer-based methods~\cite{tsai19, lv21, Liang_2021_ICCV}, indicating the importance of using rich language modality features in MER. 
Generally, we observe that fewer modalities lead to decreased performance regardless of the modality type, indicating the feasibility of multimodal approach.
We also investigate the impact of modality fusion order when combined with CMD.
Our LVA setting achieves better results than other modality configurations. 
Changing the fusion order results in decreased performance, which demonstrates the optimality of our configuration in CMD.

\begin{table}[]
\small
\caption{Comparative analysis of uni-modality, bi-modality, tri-modality inputs in CMD for sentiment analysis using CMU-MOSEI dataset. V: Vision, L: Language, A: Audio.}
\label{tab:modality_orders}
\centering
\begin{tabular}{l|cccc}
\hline
\multicolumn{1}{c|}{Modalities} & $\text{ACC}_{2}$(\%) & F1(\%) \\ \hline \hline
V                      & 74.7    & 73.8    \\
A                      & 70.9    & 69.1    \\ 
L                      & 83.7    & 83.6    \\ \hline
VA                     & 74.6    & 74.4    \\
AV                     & 75.0    & 73.9    \\
AL                     & 84.4    & 84.4    \\
LA                     & 84.5    & 84.6    \\
VL                     & 84.7    & 84.5    \\
LV                     & 84.4    & 84.4    \\ \hline
VAL                    & 84.5    & 84.3    \\
AVL                    & 84.8    & 84.7    \\
ALV                    & 84.6    & 84.5    \\
LAV                    & 85.0    & 84.7    \\
VLA                    & 85.0    & 84.8    \\
LVA (Ours)             & \textbf{85.3}    & \textbf{85.1}    \\ \hline
\end{tabular}
\end{table}

\begin{figure}[tb]
 \centering
 \includegraphics[height=3.3cm]{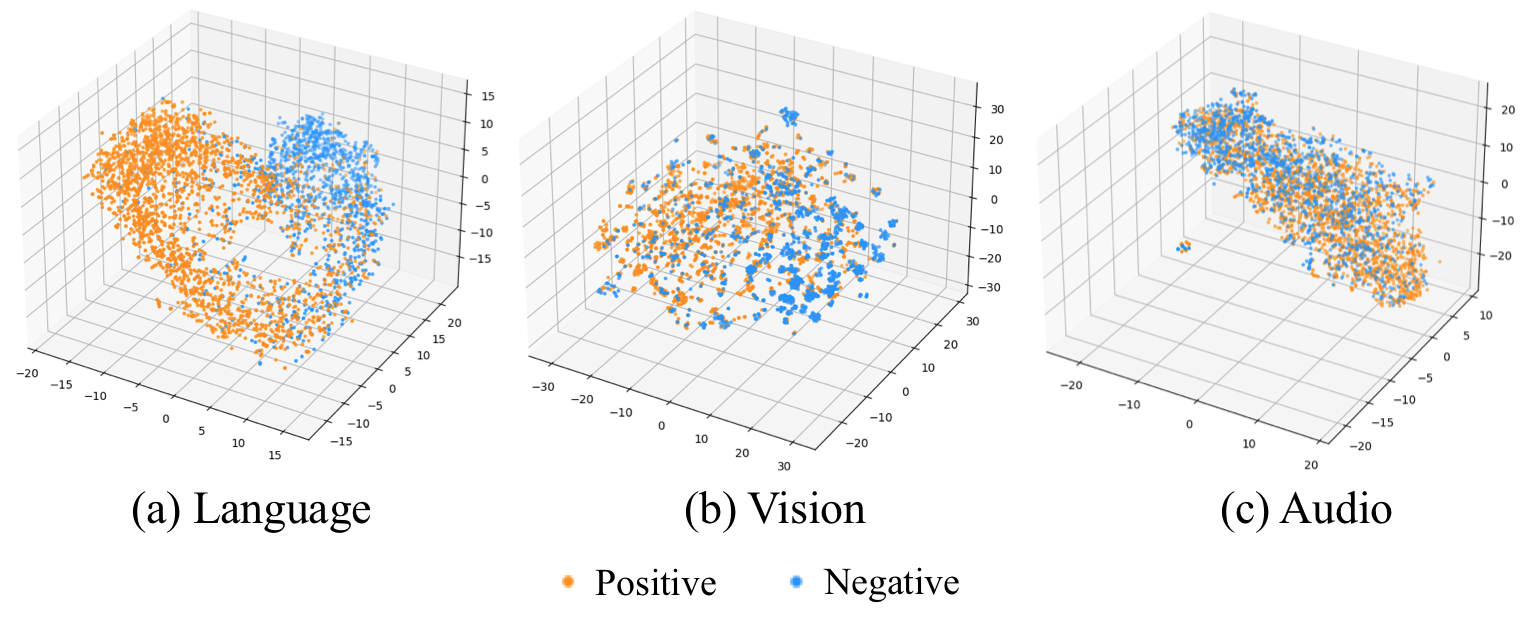}
 \caption{Visualization of the modality embeddings for sentiment analysis on CMU-MOSEI dataset.} 
 \label{fig:visualize1}
\end{figure}

We visualize the modality embeddings extracted from the modality encoders in Fig.~\ref{fig:visualize1}. The language modality provides greater separability between the \emph{positive} and \emph{negative} categories, whereas the embeddings from the audio modality are less distinguishable. This indicates that the language modality plays the most critical role in MER problems and validates our language-vision-audio (LVA) setting in CMD. 

\subsubsection{Effect of CMD and LE}
Table~\ref{tab:ablations} shows the ablation experiments of CMD and LE on CMU-MOSEI dataset. 
MER-CLIP (w/o CMD, LE) denotes MER-CLIP without both CMD and LE, MER-CLIP (w/o CMD) means MER-CLIP without CMD, and MER-CLIP (w/o LE) means MER-CLIP without LE. 

In MER-CLIP (w/o CMD, LE), we concatenated all three modalities and processed them through fully connected layers. In MER-CLIP (w/o CMD), we generated a single embedding vector through the fully connected layer and made predictions by calculating cosine similarities with label embeddings. In MER-CLIP (w/o LE), we processed the embedding vector through CMD and computed the final logits using a fully connected layer.

First, MER-CLIP (w/o CMD, LE) results in poor performance. Interestingly, MER-CLIP with either CMD or LE alone achieves more significant improvements compared to MER-CLIP (w/o CMD, LE), indicating the effectiveness of our CMD and LE components.  
Our MER-CLIP, which includes both CMD and LE, achieves greater improvements.
This indicates that using only one of the two components (CMD, LE) is insufficient for learning  embedding alignment.
Furthermore, it validates that our LE-guided prediction scheme, which incorporates both CMD and LE, enhances discriminative representation learning.

\begin{table}[]
\small
\caption{Ablation experiments on our MER-CLIP on CMU-MOSEI dataset. ``w/'' denotes our model with the component included and ``w/o'' denotes our model without the component.}
\label{tab:ablations}
\centering
\begin{tabular}{l|cccc}
\hline
\multicolumn{1}{c|}{Combinations} & $\text{ACC}_{2}$(\%) & F1(\%) \\ \hline \hline
w/o CMD, LE                 & 81.1    & 80.5    \\
w/o CMD | w LE              & 85.1    & 84.7    \\
w CMD | w/o LE              & 84.4    & 84.1    \\ 
MER-CLIP (Ours)              & \textbf{85.3}    & \textbf{85.1}    \\ \hline
\end{tabular}
\end{table}

\begin{figure}[tb]
 \centering
 \includegraphics[height=7.3cm]{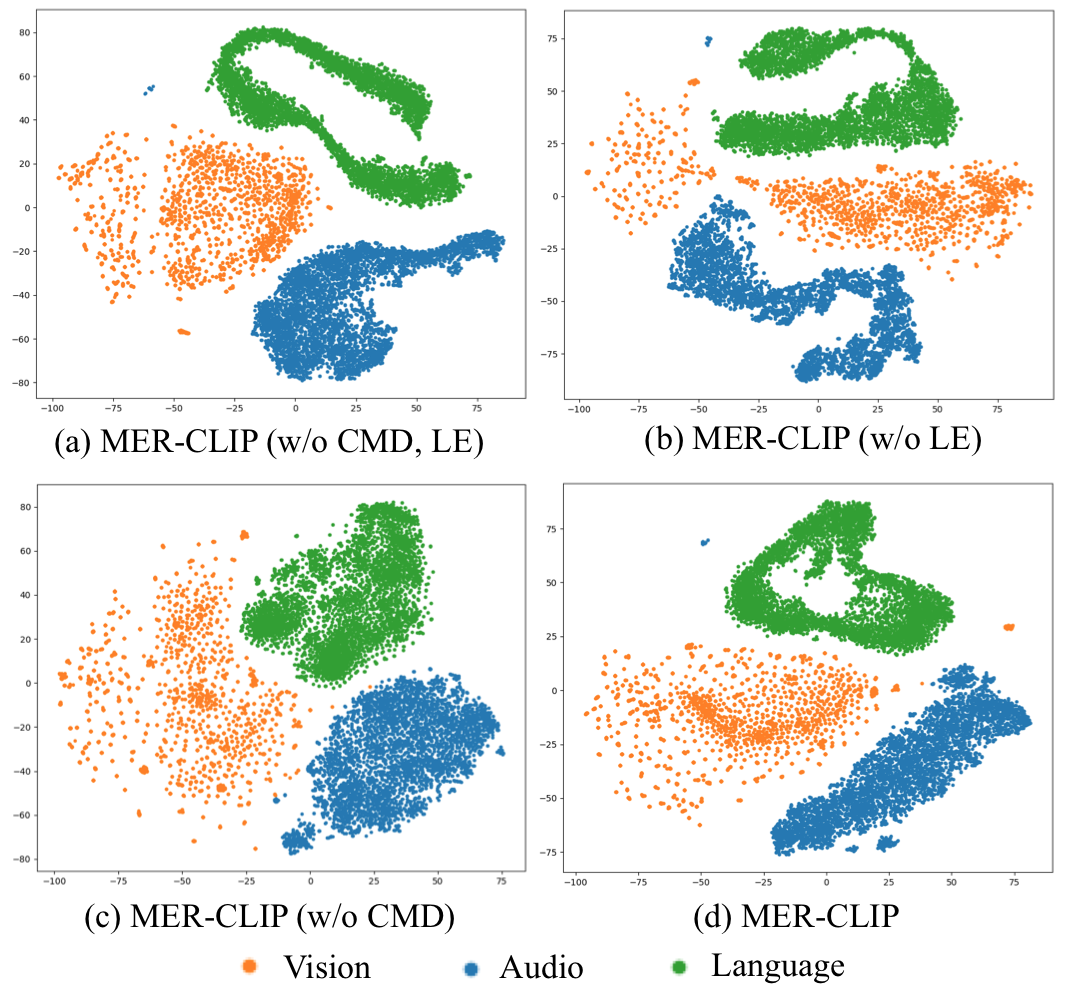}
 \caption{t-SNE visualization of the output embeddings for sentiment analysis on CMU-MOSEI dataset.} 
 \label{fig:visualize2}
\end{figure}

We visualize the output embeddings of MER-CLIP, MER-CLIP (w/o CMD, LE), MER-CLIP (w/o LE), MER-CLIP (w/o CMD) in Fig.~\ref{fig:visualize2}. 
To visualize the embeddings, we use samples in the test set of the CMU-MOSEI dataset. The embeddings of the samples are projected into a 2D space by t-SNE. 
Although the performance of MER-CLIP (w/o CMD) or MER-CLIP (w/o LE) showed a slight decrease compared to our MER-CLIP, as shown in Table~\ref{tab:ablations}, we observe a significant gap in the t-SNE visualization of the output embeddings. Combining both CMD and LE increases embedding separability between modalities, verifying MER-CLIP enhances the complementarity of heterogeneous multiple modalities.

\subsection{Generalization to different types of labels}
\label{sec:generalization}

To show the generalizability of LE, we propose using phrase-level and sentence-level labels as input to the LE.
Instead of manually creating these labels, we use a ChatGPT~\footnote{\url{https://chat.openai.com}} to automatically generate descriptions based on contextual information.
For emotion recognition, we give prompts as \texttt{``Describe the word $\left \{\texttt{emotion label} \right \}$ in one sentence without using the word $\left \{\texttt{emotion label} \right \}$''}.  
For sentiment analysis, we provide prompts as \texttt{``Describe $\left \{\texttt{sentiment label} \right \}$ sentiment without using the word $\left \{\texttt{sentiment label} \right \}$''}. 

\begin{table}[]
\caption{Sentence descriptions for emotion labels on CMU-MOSEI dataset.}
\label{tab:sentence_labels}
\centering
\begin{tabular}{l|p{6.5cm}}
\hline
\multicolumn{1}{c|}{Labels} & \multicolumn{1}{c}{Descriptions} \\ \hline \hline
Happy                           & A state of feeling delighted, content, and fulfilled, often accompanied by a bright expression and a sense of inner peace.    \\ \hline
Sad                           & A feeling of deep sorrow or unhappiness, often resulting from loss, disappointment, or misfortune.    \\ \hline
Angry                           & A strong feeling of annoyance, displeasure, or hostility that often leads to impulsive reactions or confrontations.    \\ \hline
Surprise                           & An unexpected event or piece of information that causes a strong emotional reaction, such as astonishment or wonder.    \\ \hline
Disgust                           & A profound sense of revulsion or profound disapproval elicited by something unpleasant or offensive.    \\ \hline
Fear                           & An intense sensation of dread or apprehension about a potential or imagined danger or threat.    \\ \hline
\end{tabular}
\end{table}

\begin{table}[]
\caption{Phrase descriptions for sentiment labels on CMU-MOSI and CMU-MOSEI datasets.}
\label{tab:Phrase_labels}
\centering
\begin{tabular}{l|p{6.5cm}}
\hline
\multicolumn{1}{c|}{Labels} & \multicolumn{1}{c}{Descriptions} \\ \hline \hline
Positive                           & Feelings of satisfaction, approval, or optimism.    \\ \hline
Negative                           & Feelings of dissatisfaction, disapproval, or pessimism.    \\ \hline
\end{tabular}
\end{table}

Table~\ref{tab:sentence_labels} and Table~\ref{tab:Phrase_labels} present the generated sentence and phrase descriptions for emotion and sentiment labels, respectively.
As shown in Table~\ref{tab:generalization_emotion} and Table~\ref{tab:generalization_sentiment}, using sentence or phrase labels achieves comparable performance to using word labels on accuracy and micro-F1.
It indicates generalizability to different types of labels and applicability to labels that describe the meaning of classes in the form of phrases or sentences, without requiring the inclusion of explicit class names.

\begin{table}[]
\setlength{\tabcolsep}{0pt}
\small
\caption{Performance of CMU-MOSEI dataset using sentence-level emotion labels.}
\label{tab:generalization_emotion}
\centering
\begin{tabular}{l|cccc}
\hline
\multicolumn{1}{c|}{Methods} & Accuracy(\%) & Precision(\%) & Recall(\%) & Micro-F1(\%) \\ \hline \hline
Word           & \textbf{49.3}    & 53.1     & \textbf{63.4}  & \textbf{57.8}    \\
Sentence       & 49.1    & \textbf{54.8}     & 59.2  & 56.9    \\ \hline
\end{tabular}
\end{table}

\begin{table}[]
\small
\caption{Performance of CMU-MOSI and CMU-MOSEI datasets using phrase-level sentiment labels.}
\label{tab:generalization_sentiment}
\centering
\begin{tabular}{l|cccc}
\hline
 & \multicolumn{2}{c}{CMU-MOSI} & \multicolumn{2}{c}{CMU-MOSEI} \\
\multicolumn{1}{c|}{Methods} & $\text{ACC}_{2}$(\%) & F1(\%) & $\text{ACC}_{2}$(\%) & F1(\%) \\ \hline \hline
Word                                & \textbf{84.0}    & \textbf{84.0} & \textbf{85.3}    & \textbf{85.1}   \\
Phrase                              & \textbf{84.0}    & 83.8 & 84.7    & 84.5   \\ \hline
\end{tabular}
\end{table}

\section{Conclusion}
In this paper, we propose a label encoder-guided multi-modal representation learning framework (MER-CLIP) for MER. Our framework is inspired by the observation that multimodal learning with limited and uncertain datasets imposes constraints on performance improvement. Therefore, enhanced performance in MER can be achieved by leveraging richer semantics extracted from large-scale pretrained models. 
To ensure consistent and efficient training, our framework utilizes the same CLIP architecture for both modality encoders and the label encoder. Our label encoder-guided prediction scheme enhances multimodal representation. It allows for generalizability and adaptability across diverse labels by leveraging emotion-aligned representations derived from embeddings of both the cross-modal decoder and label encoder. Experimental results demonstrate the effectiveness and applicability of the proposed method to MER.

{\small
\bibliographystyle{ieee_fullname}
\bibliography{egbib}
}

\clearpage
\twocolumn[{
\centering
{\bfseries\Large Leveraging CLIP Encoder for Multimodal Emotion Recognition}\par
\vspace{1em}
\centering
{\Large Supplementary Material}\par
\vspace{1em}
}]
\setcounter{section}{0}

In this supplementary material, we present additional experimental results in Sec.~\ref{sec:emotion_query} to show the performance of different query settings for sentiment and emotion recognition.
Examples of qualitative results are provided in Sec.~\ref{sec:demo}, and a pseudo-code for the inference process of our MER-CLIP is provided in Sec.~\ref{sec:algo}.
\section{Additional results based on different emotion queries}
\label{sec:emotion_query}

In our main paper, we use the term `\textit{Emotion}' for the emotion query in the emotion recognition task and `\textit{Sentiment}' for the sentiment analysis task, selecting terminology that aligns with the respective task names. 
To demonstrate the diverse range of words associated with emotions, we conduct experiments to compare the performance of synonyms for `\textit{Emotion}' and `\textit{Sentiment}'.

Table~\ref{tab:cmu_result} and Table~\ref{tab:cmu_result2} present comparisons of the emotion recognition and the sentiment analysis tasks, respectively, using various terms for emotion queries: `\textit{Emotion}', `\textit{Sentiment}', `\textit{Feeling}', `\textit{Impression}', `\textit{Mood}', and `\textit{Sensation}'. The term `\textit{(no word)}' indicates the exclusive use of learnable prompts for the emotion query. 
The results in Table~\ref{tab:cmu_result} indicate that `\textit{Emotion}' achieves the highest micro-F1 score, `\textit{Feeling}' achieves the highest recall score, and `\textit{Sensation}' achieves the highest accuracy/precision scores. 
Interestingly, `\textit{(no word)}' shows the comparable results across all metrics.
The results in Table~\ref{tab:cmu_result2} indicate that `\textit{Impression}' and `\textit{Mood} yield higher performance than other words, while the results in Table~\ref{tab:cmu_result3} show that `\textit{Sensation}' achieves the highest performance. 
Note that `\textit{(no word)}' also shows the comparable results across all metrics on both CMU-MOSEI and CMU-MOSI datasets.
These results indicate that the semantic information conveyed by the emotion queries impacts the performance of MER-CLIP.
We can also observe that using only learnable parameters (\textit{e.g.}, `\textit{(no word)}') without prior knowledge yields promising results, suggesting that `\textit{(no word)}' can be applied to arbitrary tasks where determining the optimal query setting is challenging.

\begin{table}[]
\setlength{\tabcolsep}{1pt}
\small
\caption{Performance comparison of words for emotion query on multimodal emotion recognition on CMU-MOSEI dataset.}
\label{tab:cmu_result}
\centering
\begin{tabular}{l|cccc}
\hline
\multicolumn{1}{c|}{Words} & Accuracy(\%) & Precision(\%) & Recall(\%) & Micro-F1(\%) \\ \hline \hline
Emotion             & 49.3    & 53.1     & 63.4  & \textbf{57.8}    \\
Sentiment           & 49.3    & 55.1     & 59.5  & 57.2    \\
Feeling             & 49.3    & 52.3     & \textbf{64.3}  & 57.7    \\
Impression          & 48.9    & 54.7     & 59.0  & 56.8    \\
Mood                & 49.2    & 53.3     & 61.9  & 57.3    \\
Sensation           & \textbf{49.6}    & \textbf{55.5}     & 59.6  & 57.5    \\
(no word)           & 49.1    & 52.7     & 62.8  & 57.3    \\ \hline
\end{tabular}
\end{table}

\begin{table}[]
\caption{Performance comparison of words for emotion query on multimodal sentiment analysis on CMU-MOSEI dataset.}
\label{tab:cmu_result2}
\centering
\begin{tabular}{l|cccc}
\hline
\multicolumn{1}{c|}{Words} & $\text{ACC}_{2}$(\%)  & F1(\%) \\ \hline \hline
Emotion                      & 85.2    & 85.1    \\
Sentiment                    & 85.3    & 85.1    \\
Feeling                      & 85.4    & 85.2    \\
Impression                   & \textbf{85.5}    & \textbf{85.5}    \\
Mood                         & \textbf{85.5}    & 85.3    \\ 
Sensation                    & 85.0    & 84.8    \\
(no word)                    & 85.0    & 85.0    \\ \hline
\end{tabular}
\end{table}

\begin{table}[]
\caption{Performance comparison of words for emotion query on multimodal sentiment analysis on CMU-MOSI dataset.}
\label{tab:cmu_result3}
\centering
\begin{tabular}{l|cccc}
\hline
\multicolumn{1}{c|}{Words} & $\text{ACC}_{2}$(\%)  & F1(\%) \\ \hline \hline
Emotion                      & 85.1    & 85.0    \\
Sentiment                    & 84.0    & 84.0    \\
Feeling                      & 83.5    & 83.4    \\
Impression                   & 84.5    & 84.3    \\
Mood                         & 83.9    & 83.7    \\ 
Sensation                    & \textbf{85.7}    & \textbf{85.5}    \\
(no word)                    & 84.0    & 84.0    \\ \hline
\end{tabular}
\end{table}

\section{Qualitative results}
\label{sec:demo}
\begin{figure*}
 \centering
 \includegraphics[height=20cm]{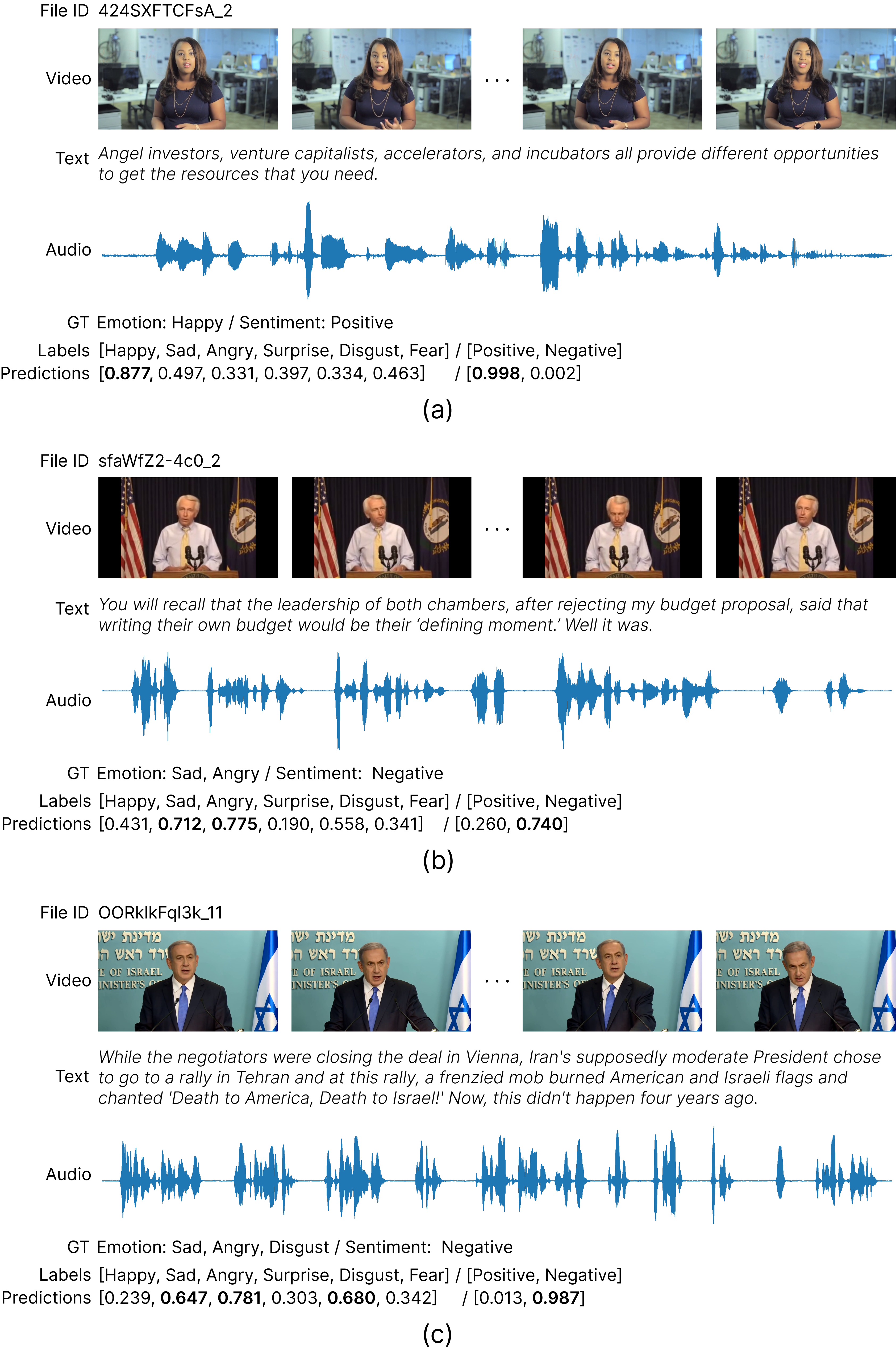}
 \caption{Qualitative results on CMU-MOSEI dataset.} 
 \label{fig:cmu_mosei}
\end{figure*}

\begin{figure*}
 \centering
 \includegraphics[height=20cm]{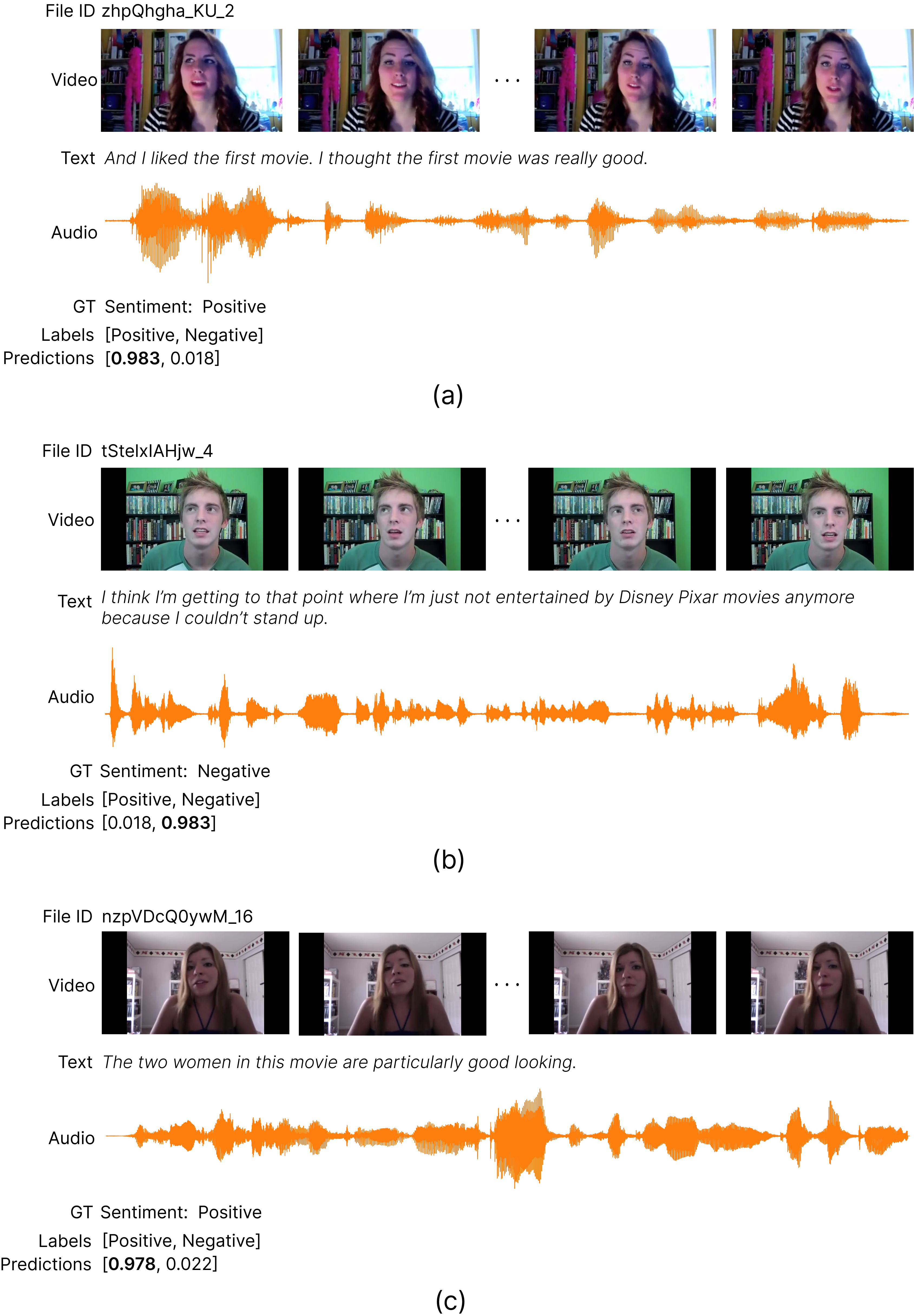}
 \caption{Qualitative results on CMU-MOSI dataset.} 
 \label{fig:cmu_mosi}
\end{figure*}

We show qualitative results for three examples, each from CMU-MOSEI and CMU-MOSI datasets, respectively.
Fig.~\ref{fig:cmu_mosei} shows results on CMU-MOSEI dataset.
The first row is the file ID, followed by three rows of input modality data.
The fifth row indicates the ground-truths for emotion recognition and sentiment analysis, and the last row shows the prediction results.
In the prediction results, probabilities exceeding the threshold (0.6) for emotion recognition are highlighted in bold to demonstrate the multi-label prediction capability, while higher scores for sentiment analysis are also highlighted in bold.
The prediction results of sentiment analysis are processed with the softmax function to clarify the results.
We can observe that our method accurately predicts labels for both emotion recognition and sentiment analysis.
Fig.~\ref{fig:cmu_mosi} shows results on CMU-MOSI dataset.
Since CMU-MOSI has only sentiment labels, we denoted the results of its predictions on sentiment analysis.
The higher score is highlighted in bold, and we can also observe that our MER-CLIP accurately predicts sentiment labels for all three examples in the CMU-MOSI dataset.

\section{Algorithms}
\label{sec:algo}
Algorithm~\ref{algo:1} presents the pseudo-code for the inference procedure of our emotion recognition and sentiment analysis tasks.

\begin{algorithm}
\caption{Inference Process for Emotion Recognition and Sentiment Analysis}
\label{algo:1}
\begin{algorithmic}[1]
\REQUIRE Visual feature $X^V$, Audio feature $X^A$, Language feature $X^L$, Emotion query embedding $Z^E$, Label embedding $Z^L$
\ENSURE Predicted class label $\hat{Y}$
\STATE Put $X^L$, $X^V$, $X^A$ in a list $M = [X^L, X^V, X^A]$ as the predetermined LVA order.
\STATE Put $M$ as key, value, and $Z^E$ as query in CMD and get the final output $Z^{[3]}$.
\STATE After processing CMD, get cosine similarity $sim(\cdot) = Z^{[3]} \cdot Z^L$.
\IF{emotion recognition}
    \STATE Apply standard normalization and sigmoid function to $sim(\cdot)$.
    \STATE Transform logits to a vector by converting values greater than the threshold (0.6) to 1 and all others to 0.
\ELSIF{sentiment analysis} 
    \STATE Multiply $sim(\cdot)$ with the learnable logit scale initialized with $exp(log(1 / 0.07))$.
    \STATE Transform logits into a vector by setting the class with the larger logit to 1 and the other class to 0.
\ENDIF
\RETURN $\hat{Y}$ \COMMENT{Return the predicted class label}
\end{algorithmic}
\end{algorithm}

\end{document}